\documentclass[10pt,twocolumn,letterpaper]{article}
\usepackage[utf8]{inputenc}
\usepackage{wacv}
\usepackage{times}
\usepackage{epsfig}
\usepackage{graphicx}
\usepackage{amsmath}
\usepackage{amssymb}
\usepackage{float}

\def\wacvPaperID{****} 

\wacvfinalcopy 

\ifwacvfinal
\def\assignedStartPage{9876} 
\fi

\ifwacvfinal
\usepackage[breaklinks=true,bookmarks=false]{hyperref}
\else
\usepackage[pagebackref=true,breaklinks=true,colorlinks,bookmarks=false]{hyperref}
\fi

\ifwacvfinal
\else
\fi

\DeclareUnicodeCharacter{2212}{-}
\begin{document}

\title{Improving Training of Text-to-image Model Using Mode-seeking Function}

\author{Naitik Bhise\\
Concordia University\\
Montreal, Canada\\
{\tt\small n\_bhise@encs.concordia.ca}
\and
Zhenfei Zhang \\
Concordia University\\
Montreal, Canada\\
{\tt\small z\_zhenfe@live.concordia.ca}
\and
Tien D. Bui \\
Concordia University\\
Montreal, Canada\\
{\tt\small bui@encs.concordia.ca}
}

\maketitle
\begin{abstract}
   Generative Adversarial Networks (GANs) have long been used to understand the semantic relationship between the text and image. However, there are problems with mode collapsing in the image generation that causes some preferred output modes. Our aim is to improve the training of the network by using a specialized mode-seeking loss function to avoid this issue. In the text to image synthesis, our loss function differentiates two points in latent space for the generation of distinct images. We validate our model on the Caltech Birds (CUB) dataset and the Microsoft COCO dataset by changing the intensity of the loss function during the training. Experimental results demonstrate that our model works very well compared to some state-of-the-art approaches.

\end{abstract}

\section{Introduction}
Text to Image synthesis is a burgeoning field and has gained a lot of attention in the last few years. With the help of Generative Adversarial Networks\cite{goodfellow2014generative}, the focus has been shifted to generate better quality images. GANs have provided the best descriptions by trying to find the relationship between the text vectors and the image. There have been techniques for the improvement of images based on the types of architectures used. Single stage methods\cite{reed2016generative} try to insert a relation between the text and the image features. Multistage methods\cite{zhang2018stackgan++}\cite{xu2018attngan}\cite{zhang2017stackgan}\cite{zhu2019dm} use the image results from the single stage to improve its image quality by applying techniques based on original image and the available word descriptions.

Though multistage architectures improve the quality of the images , there are some problems associated with them. The output images are highly dependent on the quality of the input images generated by the single stage. Also, each word in a sentence has a different level of importance\cite{xu2018attngan} on the image features. There have been architectures to address these problems such as Dynamic Memory GAN\cite{zhu2019dm} and AttnGAN\cite{xu2018attngan} that improve the semantic knowledge of the image. 

Conditional Generative Adversarial Networks(cGANs) \cite{mirza2014conditional} propose to do image generation with the help of the input descriptions and the latent vector containing the information of the input context. Therefore , they take conditional information as input augmented with the random vector for generating the output. For image synthesis, cGANs can be used in many tasks with different conditional contexts. cGANs can be applied to categorical image generation with the help of class labels. By using text sentences as contexts, cGANs can be used in text-to-image synthesis. cGANs are multimodal as they map one input to many output modes. Different output images are generated by the variation in the noise vectors that could be checked by applying cGAN on a context vector augmented with different noise vectors. One good example of variation  is the dog to cat method where the noise vectors are responsible for the change from image  of one animal to another. However, cGANs suffer from the mode collapse issue as some of the output modes are chosen as compared to others\cite{mao2019mode}. Some weak modes are suppressed as compared to the other prominent strong modes. The noise vectors are ignored or of minor impacts, since cGANs pay more attention to learn from the high-dimensional and structured conditional contexts. There are two approaches for solving the mode collapse problem. The first one focuses on discriminators by introducing different divergence metrics  and optimization process while the second deals with the auxiliary networks such as multiple generators and additional encoders. Mode seeking loss function\cite{mao2019mode} maximizes the difference between the images, enhances the modes and separates them in a simple way compared to conventional methods.

We introduce a method of implementing the Dynamic Memory-GAN\cite{zhu2019dm} architecture with the help of the mode seeking loss function. For an ensemble of points on the latent distribution, two points are chosen with different random vector and fed to the generator network. We obtain a ratio of the distance between the generated images and two noise vectors applied to the initial stage of the generator network. The discriminator is trained with the gradients from these modes so that they will not get ignored. The training is used to maximize the distance between the two random generated modes, thus exploring the target distribution and encouraging other modes.

We validate our method on two public datasets - Caltech Birds(CUB) Dataset\cite{wah2011caltech} and Microsoft COCO Dataset\cite{lin2014microsoft}. The validation is tested with the metric of the Frechet Inception Distance. From this study, an effective way to train the GANs is proposed.
The contributions of our work include:
\begin{itemize}
    \item We analyse the limitations of current techniques. In order to solve these issues, a new loss function is used to improve the training of network.
    \item Our method is effective with low parameter overheads.
    \item Our work achieves better performance on both Caltech Birds(CUB) dataset and Microsoft COCO dataset than previous models.
\end{itemize}

The review of related works to the text to image synthesis is given 
in the next section. The Procedure section entails the detailed information describing the method and it's application. The result section shows the observations with the metric values with the explanations and the conclusion gives an inference about the research.

\section{Relevant Text}
There are many methods of image generation adopted from VAEs\cite{kingma2013auto} and GANs\cite{goodfellow2014generative}. However, VAE has some disadvantages as it uses the mean square error between generated image and initial image. Compared with GAN which learns through confrontation, the image from VAE is much more blurry. DC-GAN\cite{reed2016generative} is the first work to show that cGANs can be used to generate visually acceptable images from descriptions. After DC-GAN, Generative Adversarial Network has been widely used on text to image tasks. Reed\cite{reed2016generative} introduced the single stage generation of the text to image synthesis wherein they learned the joint representation between the natural language and the real-world images and passed it to a generator for image generation. Since previous generative models failed to add the location information, Reed proposed GAWWN \cite{reed2016learning} to encode localization constraints. To diversify the generated images, the discriminator of TAC-GAN\cite{dash2017tac} not only distinguishes real images from synthetic images, but also classifies synthetic images into classes. Similar to TAC-GAN, PPGN\cite{nguyen2017plug} includes a conditional network to synthesize images conditioned on a caption. 

The limitation of the above techniques is that the generation can only understand the meaning of text except for necessary details. Moreover, it is very hard to obtain a high resolution and realistic image. Single-stage methods have spurred a series of new research techniques in this field with the multi-stage methods working towards the improvement of the image quality and the image resolution. Reed et. al. 2016 \cite{reed2016generative} generated images with the resolution of 64$\times$64 while the multistage methods took the image resolution to higher levels up to 225$\times$225. Multistage methods such as Stack-GAN\cite{zhang2017stackgan} and Stack-GAN++\cite{zhang2018stackgan++} with double and triple stages respectively.  Stack-GAN divided a complex problem (generating high-quality images) into some sub-stages that are much easier to control. 225$\times$225 images with realistic details are obtained by Stack-GAN. However, Stack-GAN does not use end-to-end training and the input is just global sentence vectors which might loss word-level information. In order to address the issues of Stack-GAN, AttnGAN\cite{xu2018attngan} was proposed to improve the multistage method by introducing an attention network between the word features and the generated image features. AttnGAN is an end-to-end model with sentence-level fine-grained information. Although AttnGAN has better inception score, it still has some limitations. For the complex scenes with multiple interactive objects, it is very difficult to understand these semantic information by using AttnGAN. The Dynamic Memory GAN\cite{zhu2019dm} uses the memory networks\cite{gulcehre2018dynamic} as the module for the semantic relationship between the image vectors and the word vectors which can solve the issue of AttnGAN effectively. The OP-GAN model\cite{hinz2019semantic} architecture has object pathways for generator and discriminator and they used a newly defined metric Semantic Object Attention(SOA) for evaluating the images produced from text descriptions. We try to use the DMGAN architecture with the best metric values on the FID metric. 

Conditional Generative adversarial networks\cite{mirza2014conditional} have been widely
used for image synthesis. With adversarial training, generators are encouraged to capture the distribution of real images. On the basis of GANs, conditional GANs synthesize
images based on various contexts.  For instances, cGANs
can generate high-resolution images conditioned on low resolution images\cite{ledig2017photo} , translate images between different
visual domains [\cite{huang2018multimodal}, \cite{isola2017image}, \cite{lee2018diverse}, \cite{liu2017unsupervised}, \cite{zhang2018unreasonable}, \cite{zhu2017toward}], generate images
with desired style \cite{li2016precomputed}, and synthesize images according to
sentences [\cite{reed2016generative}, \cite{yang2019diversity}]. Although cGANs have achieved success
in various applications, existing approaches suffer from the
mode collapse problem\cite{goodfellow2014generative}\cite{salimans2016improved}. 

Some methods focus on the
discriminator with different optimization process\cite{metz2016unrolled} and
divergence metrics [\cite{martin2017wasserstein}, \cite{mao2017least}] to stabilize the training process.
The minibatch discrimination scheme \cite{salimans2016improved} allows the discriminator to discriminate between whole mini-batches of
samples instead of individual samples.The other methods use auxiliary networks to alleviate the mode collapse issue. ModeGAN \cite{che2016mode} and VEEGAN \cite{srivastava2017veegan} enforce the bijection mapping between the input noise vectors and generated images with additional encoder networks. However, these approaches either entail
heavy computational overheads or require modifications of
the network structure, and may not be easily applicable to
cGANs. Mode seeking GAN\cite{mao2019mode} proposed a simple regularization loss function to address this issue that is dependent on the distance between the two images generated from two different points on the latent distribution. In this work, we try to achieve better image results by introducing a new loss function for the training of the Dynamic Memory GAN.

\section{Procedure}

 Generative Adversarial Networks involve solving of a Min-Max problem. The Generator generates an image based on an input distribution while the Discriminator tries to compare the generated image to the real images. Thus adversarially, the Discriminator tries to train generator with the help of its gradients and the Generator tries to fool the Discriminator with the synthesis of the realistic images. Conditional GAN generates its output distribution with the help of a directed input distribution based on a prior condition or context. Text to Image synthesis involves the use of more than a single GAN network for refining of the images called stages. We try to make use of the initial network of the DMGAN.

The DMGAN is a network that learns the inner semantic relationship between the text and the image with the help of a memory architecture. The objective of the Dynamic memory is to fuse the image and text information through non-trivia transformations between the key and value memory. It tries to improve the image vectors with the help of the text vectors. The memory is an ensemble of three modules namely the memory writing, key-value memories and the response gate. 
\begin{itemize}
    \item Memory writing: this module encodes prior information from the image vectors and the text vectors through convolutional operation. That is, obtain attention between feature map and word vector.
    \item Key addressing: In Key addressing section, key memory are used to search related memories by utilizing similarity function.
    \item Value reading: Value reading is the process of leveraging the key memories to prepare the output memory representation.
    \item Response gate: Response gate is the section for the concatenation of the image and formed vectors.
\end{itemize}
The encoder by Reed et. al. 2016\cite{reed2016learning} is used for the generation of the word embeddings and the sentence embeddings. The text encoder is a pretrained bidirectional LSTM network. The deep conventional generator predicts an initial image $x_0$ with a rough shape and few details in
accordance with the sentence feature and a random noise vector.

At the dynamic memory based image refinement stage, more fine-grained visual contents are added to the fuzzy initial images to generate a photo-realistic image $x_i: x_i = G_i(R_i−1, W)$, where $R_i−1$ is the image feature from the last stage. The refinement stage is conducted many attempts to retrieve more pertinent information and generate a high-resolution image with more fine-grained details. Figure 1 shows the diagram of the proposed method.

\begin{figure*}
    \centering
    \includegraphics[width = \textwidth]{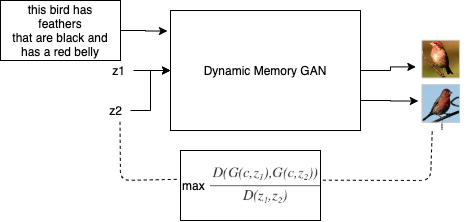}
    \caption{Architecture of the proposed method, which can solve the model collapse effectively by using mode seeking loss function.}
    \label{fig:my_label}
\end{figure*}

\subsection{Objective Function}
The objective function of the generator network is defined as:
\begin{equation}
L = \sum_i L_{G_i} + \lambda_1 L_{G_{CA}} + \lambda_2 L_{DAMSM} + \lambda L_{ms} 
\end{equation}

in which $\lambda_1$ and $\lambda_2$ are the corresponding weights of conditioning augmentation loss and DAMSM loss. We tune the $\lambda$ parameter as for the intensity of the Mode-seeking function. $G_0$ denotes
the generator of the initial generation stage. $G_i$ denotes the
generator of the i-th iteration of the image refinement stage.

\textbf{Adversarial Loss}: The adversarial loss for $G_i$
is defined as:
\begin{equation}
L_{G_i} = -\frac{1}{2}\left[ { E }_{ x\sim { p }_{ G_i } } log(D_i(x)) +{ E }_{ x\sim { p }_{ G_i } } log(D_i(x,s)) \right]  
\end{equation}
where the first term is the unconditional loss which makes
the generated image real as much as possible and the second
term is the conditional loss which makes the image match
the input sentence. Alternatively, the adversarial loss for
each discriminator $D_i$ is defined as:
\begin{equation}
\begin{split}
    L_{D_i} = -\frac{1}{2}[ \underbrace{E_{x\sim p_{data}}log(D_i(x)) +E_{ x\sim { p }_{ G_i } } log(1-D_i(x))}_{unconditional\quad loss} \\
    + \underbrace{E_{x\sim p_{data}}log(D_i(x,s)) +{ E }_{ x\sim { p }_{ G_i } } log(1-D_i(x,s))}_{conditional\quad loss}]
\end{split}
\end{equation}

Here, the unconditional loss is designed to distinguish the
generated image from real images. The conditional loss measures the matching context between the image vectors and word embeddings.

\textbf{Conditioning Augmentation Loss}: The Conditioning
Augmentation (CA) technique\cite{zhang2017stackgan} is proposed to augment training data and avoid overfitting by resampling the input sentence vector from an independent Gaussian distribution. Thus, the CA loss is defined as the Kullback-Leibler divergence between the standard Gaussian distribution and the Gaussian distribution of training data.
\begin{equation}
    L_{ CA }=D_{ KL }(N(\mu(s),\Sigma(s))||N(0,I))
\end{equation}

where $\mu$ and $\Sigma$ are mean and diagonal covariance matrix of the sentence feature.  are computed by fully connected layers.

\textbf{DAMSM Loss}: The DAMSM loss constructed by \cite{xu2018attngan} is used to measure the degree matching between images and text descriptions. It makes images well conditioned on the content of the text descriptions.

\textbf{Mode-seeking loss function}:  The mode collapse function is addressed with the help of a loss function given by the Mao et al 2019\cite{mao2019mode}. While extracting a distribution over a latent prior, many output modes are generated due to different initial vectors. Each mode generated due to random distribution generates a different image. Some of the modes are being preferred as compared to the other ones. The mode-seeking loss function tries to strengthen the minor modes of the distribution between the major modes.

Suppose a latent distribution is used to generate two different instances based on the same input. The first point $z_1$ causes the formation of an image mode $I_1=G(c,z_{ 1 })$ image and the second point $z_2$ leads to $I_2=G(c,z_{ 2 })$ image after application of a Generator and a context vector c. The mode seeking loss function is given by the expression
\begin{equation}
L _{ ms }=\underset{G}{max}\left[\frac { D(G(c,z_{ 1 }),G(c,z_{ 2 })) }{ D(z_{ 1 },z_{ 2 })} \right]
\end{equation}
where D( . , . ) is a distance function that calculates the magnitude of the distance between the two tensors or vectors. The numerator calculates the distance between the image vectors while the denominator is the distance between the normal vectors. The purpose of this term is to maximize the distance between the image vectors and thus separate the modes from overlapping.

\subsection{Implementation Details}
The DM-GAN network is trained according to the implementation details explained in the original paper\cite{zhu2019dm}. For text embedding, a pre-trained bidirectional LSTM text encoder is employed by Xu et al.\cite{xu2018attngan} and their parameters are fixed during training. Each word feature corresponds to the hidden states of two directions. The sentence feature is generated by concatenating the last hidden states of two directions. Primarily, the initial image generation stage synthesizes images with 64x64 resolution. Then, the dynamic memory based image refinement stage refines images to 128x128 and 256x256 resolution in the second and third stage. By default from \cite{zhu2019dm}, we set $N_w$ = 256, $N_r$ = 64 and $N_m$ = 128 to be the dimension of text, image and memory feature vectors respectively. We set the hyperparameter $\lambda_1$ = 1 and $\lambda_2$ = 5 for the CUB dataset and $\lambda_1$ = 1 and $\lambda_2$ = 50 for the COCO dataset. All networks are trained using ADAM optimizer\cite{kingma2014adam} with batch size 8 for COCO dataset and 5 for CUB, $\beta_1$ = 0.5 and $\beta_2$ = 0.999. The learning rate is set to be 0.0002. We train the DM-GAN model with 600 epochs on the CUB dataset and 120 epochs on the COCO dataset. 

\section{Results}
In this section, we evaluate the DM-GAN model quantitatively and qualitatively. The implementation of the DM-GAN model was done using the open-source Python library PyTorch\cite{paszke2017automatic}.

\textbf{Datasets}: To demonstrate the capability of our proposed method for text-to-image synthesis, we conducted experiments on the CUB\cite{wah2011caltech} and the COCO\cite{lin2014microsoft} datasets. The CUB dataset contains 200 bird categories with 11,788 images, where 150 categories with 8,855 images are employed for training while the remaining 50 categories with 2,933 images for testing. There are ten captions for each image in CUB dataset. The COCO dataset includes a training set with 80k images and a test set with 40k images. Each image in the COCO dataset has five text descriptions.

\textbf{Evaluation Metric}: We quantify the performance of the DM-GAN in terms of Frechet Inception Distance (FID). Each model generated 30,000 images conditioning on the text descriptions from the unseen test set for evaluation.

The FID\cite{heusel2017gans}  computes the Frechet distance between synthetic and real-world images based on the extracted features from a pre-trained Inception v3 network. It measures the distance between the synthetic data distribution and the real data distribution. A lower FID implies a closer distance between generated image distribution and real-world image distribution.

\begin{figure*}
    \centering
    \includegraphics[width = \textwidth]{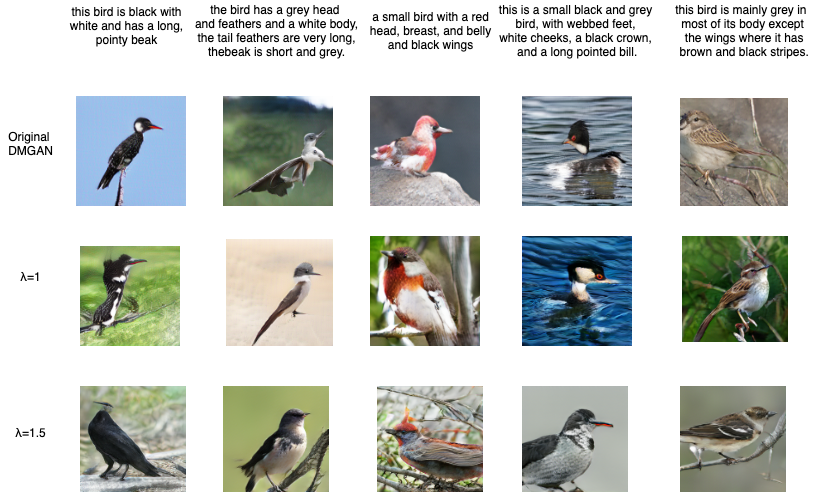}
    \caption{Results of our models on the CUB dataset with the values of the $\lambda$ coefficient and their respective images}
    \label{fig:my_label_2}
\end{figure*}
The DM GAN network was trained on two values of the $\lambda$ for each of the Datasets. The trained models were used to generate 30000 images of the validation dataset separate from the training data. The generated images were used for generating the FID metric score.
\subsection{Image Quality}
We can see that the image from the new DMGAN has better quality in terms of finer features and better characteristics of the birds for the CUB dataset than the original DMGAN images. It can also be observed that the bird is well separated from the background and there are less instances of having a double head. Also in many images, the background is clearly described and can be distinguished very well from the bird. There are few images where we can see colours well distinguishable and brighter to correlate well with the description. In the case of COCO dataset, more distinguishing features can be seen in the same way as birds with better clarity. With increasing values of parameters, we can see improvement in the image quality and better colours in the image. Thus, our training method produces better distinctive features in the images by improving contrast.

\begin{table}
\centering
\begin{tabular}{l l l}
\hline
Architecture & CUB Dataset & Coco Dataset  \\
\hline\hline
StackGAN++\cite{zhang2018stackgan++} & 35.11 & 33.88\\ \hline
AttnGAN\cite{xu2018attngan} & 23.98 & 35.49 \\
\hline
DMGAN\cite{zhu2019dm} & 16.09 & 32.64 \\
\hline
OPGAN\cite{hinz2019semantic} & - & 24.7 \\\hline
Ours @ $\lambda=1$ & 14.27 & 24.3\\
\hline
Ours @ $\lambda=1.5$& 13.91 & 26.01\\
\hline
 \end{tabular}
\caption{Evaluation results (FID score) on CUB-200-2011 and COCO dataset with two different values of $\lambda$. Ours is better than DMGAN and higher $\lambda$ has better results.}
\label{table:kysymys}
\end{table}

\subsection{Comparison with other models}
Our objective is to understand the importance of the mode seeking term in the loss function by tweaking its amplitude. It is decreased from 16.09 for the original DMGAN network to the 14.27 for the $\lambda_3=1$ while it reduced further to 13.93 for $\lambda_3=1.5$. The COCO Dataset saw a bigger change as the value of the FID score was 24.3 for the $\lambda_3=1$, a 25.5\% decrease from the original DMGAN FID and 1.6\% from the OPGAN FID. But the dataset gave a higher score for $\lambda_3=1.5$ with a value of 26.01 beating the basic model of DMGAN. It is a bigger improvement on the previous AttnGAN model with a 42\% decrease in FID on CUB dataset and 31.5\% decrease in FID on the COCO dataset. In summary , we can say that we have trained the DMGAN network to achieve the best experimental results as compared to the original loss scheme. Table \ref{table:kysymys} compiles the values of FID scores with changing parameters with the FID values of the previous architectures.

\subsection{Analysis}
We observe the variation of the FID with the increase in the $\lambda$ factor. The $\lambda$ factor is responsible for the strength of the mode seeking loss function in the overall loss expression and thus, it has an influence on the training of the network. It manages the intensity of the mode collapse loss function in the loss term. We can see in Table \ref{table:kysymys} the results for the FID scores on different values of parameters. We see that the FID score decreases as we increase the amplitude of the mode seeking term for the COCO dataset but it decreases for the CUB dataset. We have found some modes that resulted in the new images having different shades and better quality. Therefore, we have achieved a substantial amount of diversity in our network and thus we could generate better images for descriptions.

\begin{figure*}
    \centering
    \includegraphics[width = \textwidth]{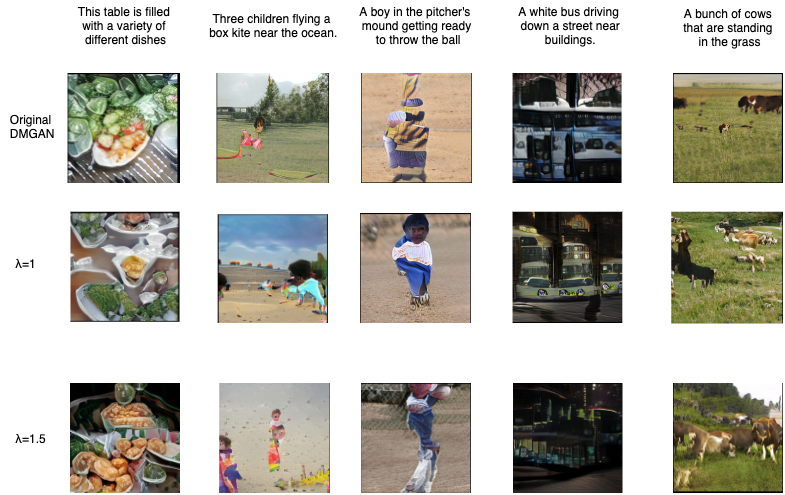}
    \caption{Results of our models on the Microsoft COCO dataset with the values of the $\lambda$ coefficient and their respective images}
    \label{fig:my_label2}
\end{figure*}

Qualitatively, we can see that the increase in $\lambda$ improves the quality of the image with the increase in clarity. But it also increases the strength of the mode-seeking term which depletes the importance of the DAMSM loss and the conditional loss in the original loss function. Some of the effects can be seen in the first example of the bird\ref{fig:my_label_2}. We can see that the bird being black is magnified better in the higher $\lambda$ and other features are rendered less significance. We can see in the fourth example of \ref{fig:my_label2} that the bus being white is being jumbled up as we increase the $\lambda$.  

\section{Conclusion}
In this work, we present a simple but effective mode seeking regularization term on the generator to address the model collapse issue in cGANs and improve the quality of the images generated by the Dynamic Memory GAN. By maximizing the distance between generated images with respect to that between the corresponding latent codes, the regularization term forces the generators to explore more minor modes. It is concluded that changing the amplitude of the regularization term could lead to better results and images. We measured the efficiency of our method with the help of the FID metric as well as human eye evaluation and the new technique was applied to two image datasets. Our model compares well with the past state-of-the-art models in terms of the FID score. Both qualitative and quantitative results show that the proposed regularization term facilitates the baseline frameworks, improving the diversity without sacrificing visual quality of the generated images.

\bibliographystyle{ieee_fullname}
\bibliography{ms}

\end{document}